\documentclass{article}
\usepackage[utf8]{inputenc}
\usepackage{amssymb}
\usepackage{amsmath}
\usepackage{graphicx}
\usepackage{listings}
\usepackage{indentfirst}
\usepackage{bbm}
\usepackage{mathtools}
\usepackage[utf8]{inputenc}
\usepackage[english]{babel}
\usepackage[margin=1in]{geometry}
\usepackage{natbib}
\usepackage[super]{nth}
\usepackage{bm}
\usepackage{verbatim}
\usepackage[dvipsnames]{xcolor}
\usepackage{hyperref}
\hypersetup{
    colorlinks=true,
    linkcolor=blue,
    filecolor=magenta,      
    urlcolor=purple,
    citecolor=NavyBlue}

\DeclareMathOperator{\ex}{\mathbb{E}}

\newcommand{\inner}[2]{\left\langle #1, #2 \right\rangle}

\newcommand{\norm}[1]{\lVert #1 \rVert}
\newcommand{\paren}[1]{\left( #1 \right)}
\newcommand{\bracket}[1]{\left[ #1 \right]}

\newcommand{\mbf}[1]{\mathbf{#1}}

\newcommand\convDist{\stackrel{\mathclap{\normalfont\mbox{\emph{d}}}}{\longrightarrow}}
\newcommand\convProb{\stackrel{\mathclap{\normalfont\mbox{\emph{p}}}}{\longrightarrow}}
\newcommand\numberthis{\addtocounter{equation}{1}\tag{\theequation}}

\title{Issues with Neural Tangent Kernel Approach to Neural Networks}
\author{
Haoran Liu\footnote{Department of Statistics, Indiana University Bloomington}, 
Anthony Tai\footnote{Naval Surface Warfare Center - Crane Division}, 
David J. Crandall\footnote{Department of Computer Science, Indiana University Bloomington}, 
Chunfeng Huang\footnotemark[1]
}
\date{}

\begin{document}





\maketitle
\begin{abstract}
    Neural tangent kernels (NTKs) have been proposed to study the behavior of trained neural networks from the perspective of Gaussian processes. An important result in this body of work is the theorem of equivalence between a trained neural network and kernel regression with the corresponding NTK. This theorem allows for an interpretation of neural networks as special cases of kernel regression. However, does this theorem of equivalence hold in practice?

    In this paper, we revisit the derivation of the NTK rigorously and conduct numerical experiments to evaluate this equivalence theorem. We observe that adding a layer to a neural network and the corresponding updated NTK do not yield matching changes in the predictor error. Furthermore, we observe that kernel regression with a Gaussian process kernel in the literature that does not account for neural network training produces prediction errors very close to that of kernel regression with NTKs. These observations suggest the equivalence theorem does not hold well in practice and puts into question whether neural tangent kernels adequately address the training process of neural networks.
    
\end{abstract}

\section{Introduction}
        Recent advancements in generative models such as ChatGPT \citep{chatgpt} and MidJourney \citep{midjourney} have sparked an uptick in discussions of AI, including areas of adoption and potential increases in efficiency. 
        However, these advances in technology are not without risks. 
        The Database of AI Litigation \citep{dail} cites a number of AI-related lawsuits covering a wide variety of issues including copyright infringement in the training data, unfair competition via coordinated rent manipulation, misrepresentation of the safety of autonomous systems, and the lack of human review in automated decision making (health insurance, fraud detection, and job interviews).

        The fundamental building block of many advanced AI systems are neural networks. At a high level, neural networks are complex nonlinear models that learn from data. They are often used for prediction, dimension reduction, or generative tasks. While neural networks have shown to perform well at certain tasks, a limitation of such models is the lack of meaningful interpretation. This makes it difficult to understand precisely why these models make the decisions that they do and makes it difficult to come up with regulations or guidelines for safe, ethical, and trustworthy implementations. New technologies do require trial and error before becoming well-refined tools. Nevertheless, the importance of theoretical understanding of these models should not be dismissed. 
    
        One popular framework for understanding the behavior of neural networks is the ``Gaussian process view'' (\citet{arora2019exact}, \citet{jacot2018neural}, \citet{lee2017deep}, \citet{neal1996priors}). This framework addresses the behavior of neural networks as the widths of hidden layers become large.
        \citet{neal1996priors} showed that a pre-trained (parameters initialized but not trained) neural network with a single hidden layer converges to a Gaussian process as the width of the hidden layer approaches infinity. 
        \citet{lee2017deep} extends the work of \citet{neal1996priors} and showed that for a pre-trained fully connected neural network with arbitrary number of hidden layers ($\geq 1$) converges to a Gaussian process as the widths of the hidden layers approach infinity. 

        \citet{jacot2018neural} developed the neural tangent kernel (NTK) to describe trained neural networks under some assumptions. The key claim is that changes to the network parameters can be described by a random kernel which converges, as training time goes to infinity, to a deterministic kernel: the neural tangent kernel \citep{jacot2018neural}.
        \citet{arora2019exact} extends the neural tangent kernels to include convolutional operations and proposed a theorem of equivalence between trained neural networks and kernel regression predictor using the corresponding NTKs. This theorem of equivalence allows one to interpret neural networks as special cases of kernel regression.
        The body of work on neural tangent kernels have been extended to recurrent neural networks \citep{alemohammad2022recurrent} and attention-based models \citep{hron2020infinite} as well.
        
    \subsection{Related Works}        
        
        There are some existing critiques of the NTK.       
        \citet{simon2022reverse} argued that increasing the layers in the NTK formula does not provide additional kernel expressivity. This runs counter to the recursive nature of the NTK formula from \citet{jacot2018neural} and \citet{arora2019exact} where the number of recursive steps depends on the number of layers for the corresponding neural network. 
        \citet{chizat2019lazy} showed through numerical experiments that the performance of convolutional neural networks (CNNs) used in practice degrade as they become ``lazy'' and hence the success of CNNs cannot be adequately explained by ``lazy training.''
        \citet{seleznova2022analyzing} showed that whether the lazy training assumption holds depends on the variance of parameter initialization and in the ``chaotic phase'' (situations when lazy training doesn't hold), the NTK matrix does not stay constant during training.
        
    \subsection{Contributions}
        We evaluate the theorem of equivalence \citep{arora2019exact} between neural networks and kernel regression with the corresponding neural tangent kernel  through numerical experiments. Our results show that kernel regression with the neural tangent kernel does not adequately address the behavior of trained neural networks. 
        Furthermore, we show that kernel regression with the Gaussian process kernels (which correspond to initialized but untrained neural networks) has predictive performance very similar to that of kernel regression using the neural tangent kernels. 
    
\section{Background knowledge and methods used}
    
    \subsection{Fully Connected Neural Networks}

    
    In general, the $i^{th}$ component of the output of a fully connected neural networks with $\ell$ hidden layers can be expressed as: 
    \begin{align*}
        & z_i^{(\ell)}(\mbf{x})
        =
        b_i^{(\ell)} 
        +
        \sum_{j_\ell = 1}^{n_\ell} 
        W_{i j_\ell}^{(\ell)}
        g_{j_\ell}^{(\ell)}(\mbf{x}) \\
        &
        g_{j_\ell}^{(\ell)}(\mbf{x}) 
        =
        \sigma
        \paren{
            z_{j_\ell}^{(\ell-1)}(\mbf{x})
        }  \label{general NN}\numberthis \\
        &
        j_\ell = 1, 2, \dots, n_\ell , \quad g_j^{(0)}(\mbf{x}) = x_j
    \end{align*}
    Where 
        \begin{itemize}
            \item $z_i^{(\ell)}(\mbf{x})$: the $i^{th}$ element of the output vector after $\ell$ hidden layers with input $\mbf{x}$ \footnote{While in general these network have vector-valued outputs, this paper specifically addresses regression problems with one-dimensional outputs}.
            \item $n_0$: dimension of the input vector. 
            \item $n_\ell$: width of the $\ell^{th}$ hidden layer.
            \item $W_{ij}^{(\ell)}$: weight that associates node $j$ in the current layer with node $i$ of the next layer. The $(\ell)$ indicates that this weight immediately proceeds the $\ell^{th}$ hidden layer. 
            \item $\sigma$: a nonlinear activation function. In this paper, we use the ReLU activation function $\sigma(x) = \max(0, x)$. 
            \item $x_k$: the $k^{th}$ element of the input vector $\mbf{x} \in \mathbb{R}^{n_0}$.
            \item $b_i^{(\ell)}$: bias term added after $\ell$ hidden layers to form node $i$ of the next layer.
        \end{itemize}        
        
    
    There is a modified version of the above setup that is necessary for neural tangent kernels:
    \begin{align*}
        z_i^{\ell}(\mbf{x})
        =
        b_i^{(\ell)} 
        +
        \frac{1}{\sqrt{n_\ell}}
        \sum_{j_\ell=1}^{n_\ell} 
        W_{i j_{\ell}}^{(\ell)}
        g_{j_{\ell}}^{(\ell)}(\mbf{x}) 
        \label{NN_NTK}\numberthis{}
    \end{align*}
    The additional scaling factors $\frac{1}{\sqrt{n_\ell}}$ are added because they allow for the use of the Weak Law of Large Numbers in the NTK derivation \citep{jacot2018neural}.

    \subsection{Connection between Neural Networks and Gaussian Processes}

    \citet{neal1996priors} showed that as the width of the hidden layer goes to infinity, the scalar output of pre-trained neural networks with one hidden layer (with independent and identically distributed (i.i.d.) parameter initialization) converges in distribution to a Gaussian process.
    This results require the following assumptions:
        (i) The weight and bias parameters are i.i.d.
        (ii) The width of hidden layers $\to \infty$.
        
    \citet{lee2017deep} extends the result of \citet{neal1996priors} and showed by induction that the scalar output of pre-trained neural networks of arbitrary depth (with i.i.d. parameter initialization) converge in distribution to Gaussian processes as the widths of the hidden layers approach infinity. 
    \begin{equation}{\label{GP converge}}
        z_{i}^{(\ell)}(\mbf{x}) \convDist GP(\pmb{0}, K^{(\ell)}(\mbf{x}, \mbf{x}'))
    \end{equation}
    Here, 
        (i) $\ell$: number of hidden layers. 
        (ii) $z_i^{(\ell)}(\mbf{x})$: $i^{th}$ component of a neural network output with $\ell$ hidden layers and input vector $\mbf{x} \in \mathbb{R}^{n_0}$.
        (iii) $K^{(\ell)}(\mbf{x},\mbf{x}')$: covariance function of the Gaussian Process above. The inputs $\mbf{x}$ and $\mbf{x}'$ are vectors in $\mathbb{R}^{n_0}$.

    These results describe the behaviors of infinitely wide pre-trained neural networks with i.i.d. parameters initializations. These results serve as the foundation for the neural tangent kernel \citep{jacot2018neural} and the equivalence theorem in \citet{arora2019exact}. 
    
    \subsection{A Brief Introduction To Neural Tangent Kernels}

     \citet{jacot2018neural} developed the neural tangent kernel to address the behavior of trained neural networks. These results require a few additional assumptions on top of those used in the pre-trained network convergence results.      
     In addition to i.i.d. parameter initialization and having the widths of hidden layers go to infinity, neural tangent kernels require
         (i) Additional scaling factors in the construction of the corresponding neural network. 
         (ii) Infinitesimal learning rate.
         (iii) Training time $\to \infty$.
         (iv) Lazy training assumption: the parameters of the neural network do not change much from their initial values.
         (v) Stochastic gradient descent as the optimizer.
         (vi) Squared error loss function.
     The lazy training assumptions allows one to take a Taylor expansion about the initial parameter values that is fundamental to the derivation of NTKs (hence the ``tangent'' in neural tangent kernel). 

    A brief overview to the concept of neural tangent kernels \citep{jacot2018neural} \citep{feizi_notes}:
    
    Let $z^{(\ell)}(\pmb{\theta}, \mathbf{x})$ denote the output of a neural network $z^{(\ell)}$ with $\ell$ hidden layers with input $\mathbf{x} \in \mathbb{R}^{n_0}$.    
        (i) $\pmb{\theta} = 
        \begin{pmatrix}
            b_1^{(1)}, 
            \{W_{1j}^{(1)}\}_{j}, 
            \{b_j^{(0)}\}_{j},
            \{W_{jk}^{(0)}\}_{jk}, 
            \dots
        \end{pmatrix}^\top$ is a vector of parameters of the neural network $z^{(\ell)}$
        (ii) $\pmb{\theta}^{(0)}$ are the initial values of parameters of the neural network
        (iii) $\mathbf{x}, \mathbf{x'}$ are input vectors
        
    If we take a first-order Taylor expansion about the initial parameters 
    $\pmb{\theta}^{(0)}:$
    \citep{feizi_notes}
    \begin{align*}
        z^{(\ell)}(\pmb{\theta},\mathbf{x}) 
        &\approx
        z^{(\ell)}(\pmb{\theta}^{(0)}, \mathbf{x}) 
        +
        \frac{\partial}{\partial \pmb{\theta}} 
        z^{(\ell)}(\pmb{\theta}^{(0)}, \mathbf{x})^\top
        (\pmb{\theta}-\pmb{\theta}^{(0)}) \\
        \phi(\mathbf{x}) &= \frac{\partial}{\partial \pmb{\theta}}
        z^{(\ell)}(\pmb{\theta}^{(0)}, \mathbf{x}) \\
        \Theta(\mathbf{x}, \mathbf{x'}) &= \langle \phi(\mathbf{x}), \phi(\mathbf{x'}) \rangle
        \numberthis
    \end{align*}
    $\Theta$ is the corresponding Neural Tangent Kernel (NTK) of the neural network $z^{(\ell)}$. 
    \subsection{The Equivalence Theorem}
        
        \citet{arora2019exact} builds on top of the results of \citet{jacot2018neural} and came up with a theorem of equivalence between neural networks and kernel regression using the corresponding neural tangent kernels. 

        This theorem requires the following assumptions on top of the ones needed to derive the NTK. 
            (i) The activation function must be ReLU.
            (ii) The widths of hidden layers should be the same and be sufficiently large.
            (iii) $\norm{\mbf{x}_{te}} = 1$, the input vectors for the test set lie on the surface of a hypersphere with radius 1.

        Let $\Theta(\mbf{x}, \mbf{x}') 
            =
            \inner
            {\frac{\partial z^{(\ell)}(\mathbf{x})}{\partial \pmb{\theta}}}
            {\frac{\partial z^{(\ell)}(\mathbf{x}')}{\partial \pmb{\theta}}}$
            
            the prediction for a new data point $\mathbf{x^{*}} \in \mathbb{R}^{p}$ is
            \begin{align*}
                \widehat{z^{(\ell)}}(\mathbf{x}^{*}) 
                &= 
                \paren{
                    \Theta\paren{\mathbf{x}^{*}, \mathbf{x}_1}, 
                    \dots ,
                    \Theta\paren{\mathbf{x}^{*}, \mathbf{x}_n}
                }
                \paren{
                    H
                    + 
                    \lambda 
                    I_n
                }^{-1}
                \mathbf{y} \\
                H_{ij} 
                &=
                \Theta(\mbf{x}_i, \mbf{x}_j)
                \label{kernel_regression}
                \numberthis
            \end{align*}
        $\widehat{z^{(\ell)}}(\cdot)$ is the kernel ridge predictor \citep{welling_kernel_ridge_regression} using the NTK that corresponds to the neural network $z^{(\ell)}$.
        $\lambda$ is the coefficient of the ridge penalty ($\lambda = 0$ implies that no ridge penalization is applied). 
        \citet{arora2019exact} showed that under the NTK assumptions, when $\lambda = 0$, $\widehat{z^{(\ell)}}(\mbf{x})$ is equivalent to $z^{(\ell)}(\mbf{x})$. 

        This theorem allows for an interpretation of neural networks: under the assumptions of this equivalence theorem, neural networks are special cases of kernel ridge regression. This result, if it can be applied in practice, is incredibly powerful. Not only will it allow for interpretation of black-box models, but it can also give insights on how to construct neural network architectures in more principled ways if it holds in practice. 
        
\section{Derivation of NTK and other models}

    \subsection{Models (Detailed derivations can be found in the appendix)}
    There are different formulas for NTK in the literature. To evaluate the equivalence theorem, we first derive the NTK with 1 and 2 hidden layers in this subsection. Our derivations are rigorous, taking bias term into account carefully with more general assumptions.  

    \textbf{[NTKB1 and NTKB2]} correspond to wide neural networks with 1 and 2 hidden layers, respectively. NTKB takes bias terms into account and contains additional constants to adjust variance. The other difference between NTKB and NTKJ (the NTK in \citet{jacot2018neural}) is the way the first layer is handled. NTKJ rescales the linear combinations $\sum_{k=1}^{n_0} W_{ik}^{(0)}x_k$ by a factor of $\frac{1}{\sqrt{n_0}}$ while keeping all parameter initializations to be $\mathcal{N}(0,1)$. NTKB does not include the scaling factor for the first layer but sets the initialization of $W_{ik}^{(0)}, k=1,2,\dots,n_0$ to be $\mathcal{N}(0, \frac{\sigma_{w,0}}{\sqrt{n_0}})$. All other parameters in NTKB are still initialized from $\mathcal{N}(0,1)$. NTKB handles the inputs this way because the scaling factors are meant to facilitate the weak law of large numbers. However, as $n_0$ is finite there is no need to have the $\frac{1}{\sqrt{n_0}}$ scaling factor. 

    Here we present the formulas for NTKB1 and NTKB2 that we derived.
\begin{equation}
    \Theta_{NTKB1}(\mbf{x},\mbf{x'})
    = 1+ 
    \frac{c_1}{2\pi} 
    \sqrt{d_{\mbf{x}} d_{\mbf{x'}}} 
    (\sin \delta^{(0)}+\cos \delta^{(0)}(\pi - \delta^{(0)}))
    + c_1 \frac{\sigma^2_{w,1}}{2 \pi} (\pi - \delta^{(0)}) + c_1 \frac{\sigma^2_{w,1}}{2\pi}\inner{\mathbf{x}}{\mathbf{x}'}(\pi - \delta^{(0)})
\end{equation}
where
\begin{align*}
    &d_{\mbf{x}} = \frac{\sigma^2_{w,0}}{d_{in}} \norm{\mbf{x}}^2+\sigma^2_{b,0}, \quad
    d_{\mbf{x'}} = \frac{\sigma^2_{w,0}}{d_{in}} \norm{\mbf{x'}}^2+\sigma^2_{b,0}, \\
     &\delta^{(0)} = \arccos\left(\frac{ \frac{\sigma^2_{w,0}}{d_{in}} \inner{\mathbf{x}}{\mathbf{x}'} +\sigma^2_{b,0}}{\sqrt{d_{\mbf{x}}d_{\mbf{x'}}}} \right),
     \numberthis
\end{align*}
and the parameters $c_1, \sigma_{w,0}, \sigma_{w,1}$ and $\sigma_{b,0}$ can be set by the user. In this paper, $c_1 = 2$ and $\sigma_{w,0} = \sigma_{w,1}=\sigma_{b,0} = 1$.

As for NTKB2, we have
\begin{equation}
    \Theta_{NTKB2}(\mbf{x},\mbf{x'})
    = 
    1
    + c_2 
    \sigma_{w,2}^2 d_1 
    + c_1 c_2 
    \sigma^2_{w,1}
    \sigma^2_{w,2}
    (1+\inner{\mathbf{x}}{\mathbf{x}'})
    d_1 d_0 
    + c_1 c_2 
    \sigma^2_{w,2}
    d_1 q_0
    + c_2 q_1,
\end{equation}

where
\begin{align*}
&d_1 = \frac{1}{2\pi} (\pi - \delta_K^{(1)}), \quad
d_0 = \frac{1}{2\pi} (\pi - \delta_K^{(0)}),\\
&q_1 = \frac{1}{2\pi} \sqrt{K^{(1)}(\mbf{x},\mbf{x})K^{(1)}(\mbf{x}',\mbf{x}')}(\sin \delta_K^{(1)}+\cos \delta_K^{(1)}(\pi - \delta_K^{(1})) \\
&q_0 = \frac{1}{2\pi} \sqrt{K^{(0)}(\mbf{x},\mbf{x})K^{(0)}(\mbf{x}',\mbf{x}')}
\times (\sin \delta_K^{(0)}+\cos \delta_K^{(0)}(\pi - \delta_K^{(0})) \\
&K^{(0)}(\mbf{x},\mbf{x}') = \frac{\sigma_{w,0}^2}{d_{in}} \inner{\mbf{x}}{\mbf{x}'} + \sigma^2_{b,0}, \quad
K^{(1)}(\mbf{x},\mbf{x}')=\sigma^2_{b,1} + c_1 \sigma^2_{w,1} q_0 \\
&\delta_K^{(0)} = \arccos \left( \frac{K^{(0)}(\mbf{x},\mbf{x}')}{\sqrt{K^{(0)}(\mbf{x},\mbf{x})K^{(0)}(\mbf{x}',\mbf{x}')}} \right), \quad
\delta_K^{(1)} = \arccos \left( \frac{K^{(1)}(\mbf{x},\mbf{x}')}{\sqrt{K^{(1)}(\mbf{x},\mbf{x})K^{(1)}(\mbf{x}',\mbf{x}')}} \right)
\end{align*}
and we set $c_2=2, \sigma_{w,2}=\sigma_{b,1}=1$ in this paper.

    \textbf{Remark: } Given NTKB1 and NTKB2, we can evaluate the equivalence theorem via aa numerical study (in Section 3.2). We will simulate the data from NTKB1. With this simulated data, we carry out the following prediction tasks for comparisons. (i) Use one and two layer NNs with wide widths; (ii) Kernel regressions with NTKs, including NTKB, NTKJ, NTKA; (iii) Kernel regression with pre-trained kernels, termed GP1 and GP2 \citep{lee2017deep}. (iv) Kernel regression with an arbitrary kernel: the Laplace exponential covariance kernel $K1(\mbf{x}, \mbf{x}') = a \exp (-\norm{\mbf{x}-\mbf{x}'}/b)$, with $a=2, b=6$. Note that this kernel is arbitrary and unrelated to neural networks. 
    
    \textbf{[NTKJ1 and NTKJ2]} correspond to wide neural networks with 1 and 2 hidden layers, respectively. This is the original NTK by \citet{jacot2018neural}. It assumes that all parameters are i.i.d $\mathcal{N}(0,1)$ and take the bias terms into account. 
        
    \textbf{[GP1 and GP2]} correspond to wide neural networks with 1 and 2 hidden layers that are initialized but not trained. This model was derived by \citet{neal1996priors} and \citet{lee2017deep} in their works on the Gaussian process view of neural networks. The general gist is that an infinitely wide neural network upon random initialization of its parameters (i.i.d.) converges in distribution to a Gaussian process as the width of the hidden layer approaches infinity. 
    \citet{lee2017deep} extended the work of \citet{neal1996priors} to an arbitrary number of hidden layers. 

    \textbf{[NTKA1 and NTKA2]} correspond to wide neural networks with 1 and 2 hidden layers, respectively. This variant of the NTK is by \citet{arora2019exact}. This model does not take the bias terms into account and has parameters initialized as i.i.d. $\mathcal{N}(0,1)$. However, unlike Jacot's original NTK, this variant does not have the $\frac{1}{n_0}$ rescaling factor.

    \textbf{[K1 (Laplacian kernel)]} This kernel is not derived from the architecture of a neural network. It serves as a ``non NN-related kernel'' for us to compare results with in the experiments.
    
    \textbf{[NN1 and NN2]} are wide fully connected neural networks of 1 and 2 layers, respectively. The width of hidden layers is set to be $10,000$ and both networks use stochastic gradient descent (SGD) as the optimizer. The output is one-dimensional. The learning rate is set to $0.002$ and the 1-layer network is trained for 3000 epochs and the 2-layer network is trained for 6000 epochs. In order to properly evaluate the equivalence claim, We chose the widths to be wide, the optimizer to be SGD, the learning rate to be small, and number of epochs to be large.
       
\section{Experiment Setup and Results}

    \subsection{Generating Simulated Data}
        In this experiment, we simulate data from NTKB1, then carry out prediction tasks listed in the Remark in the section 3.1.
        In particular, the input dimension  is set to 15 ($d_{in}=15$). The total number of observations is 1000 ($n_{obs}=1000$), which will later be split into training and test sets. The response is univariate because we want to do a simple prediction problem that uses a squared error loss function. 
        A total of 50 trials ($n_{trials}=50$) are conducted and the predictive performance is measured by root mean squared error (RMSE). 
        In each trial, a new design matrix (and response vector) is generated. The procedure for simulated data generation is as follows: 
        \begin{enumerate}
            \item We first sample $d_{in} \times n_{obs}$ scalar values from $\text{unif}[-5, 7]$. These scalar values are then molded into a design matrix $\mathbf{X}$ with dimensions $d_{in} \times n_{obs}$. Then then rows of the design matrix $\mathbf{X}$ are rescaled so that each row has unit norm.
            \item Then we compute the covariance matrix $\mbf{H}$. 
            $\mbf{H}$ is an $n_{obs} \times n_{obs}$ matrix whose $ij^{th}$ entry is  
            $\mathbf{H}_{ij}
            =
            \Theta_{NTKB,1} \paren{\mbf{x}_i, \mbf{x}_j}
            $ as in Section 3. 
            \item Once the covariance matrix has been populated, we generate a response vector (that includes both training and test observations at this point) via multivariate normal distribution. The response vector generated should be of length $n_{obs}$.
            \item Then we split the design matrix and response vector to be $\frac{1}{3}$ test and $\frac{2}{3}$ training. 
        \end{enumerate}
        We will test the predictive abilities of the models described in section 3.1 using the above simulated data set.  
    \begin{figure}[h!]
        \begin{center}
            \includegraphics[scale=0.8]{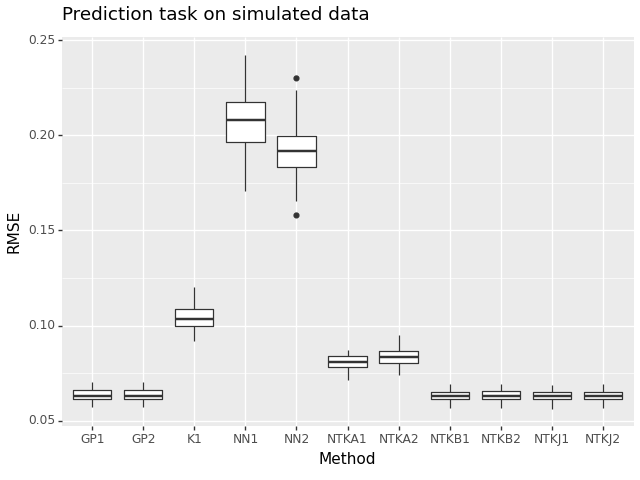}
        \end{center}
        \caption{Boxplot of RMSE over 50 trials for each model}
        \label{boxplot_results}
    \end{figure}

   \begin{figure}[h!]
        \begin{center}
            \begin{tabular}{ |c|c| } 
            \hline
            Method  & Sample Mean RMSE   \\
            \hline
            \hline
            NN (1 layer)  & 0.207151  \\ 
            \hline
            NN (2 layers)  & 0.191916  \\ 
            \hline
            GP (1 layers)  & 0.0634271  \\ 
            \hline
            GP (2 layers)  & 0.0635353  \\ 
            \hline
            NTKA (1 layers)  & 0.0808606  \\ 
            \hline
            NTKA (2 layers)  & 0.0836348  \\ 
            \hline
            NTKJ (1 layers)  & 0.063008  \\ 
            \hline
            NTKJ (2 layers)  & 0.062946  \\ 
            \hline
            NTKB (1 layers)  & 0.0629396  \\ 
            \hline
            NTKB (2 layers)  & 0.0629477  \\ 
            \hline
            Laplacian (K1)  & 0.103857  \\ 
            \hline
            \end{tabular}
        \end{center}
        \caption{Table of Mean RMSE for each model. }
        \label{table of RMSE}
    \end{figure}
    
    \subsection{Results}
        In Figure \ref{table of RMSE}, we show the average RMSEs for these predictors across 50 trials. In Figure \ref{boxplot_results}, the boxplots for the RMSEs for each of the predictors are displayed. NTKB1 has the best predictive performance, as it is the ground truth. The NTKJ and GP kernel predictors are very close in terms of mean RMSE. 
        
        These results are not consistent with the claims of equivalence. We can see in Figure \ref{boxplot_results} that the neural network gained performance with an additional layer but NTKA lost some performance with an additional layer. Furthermore, NTKB and NTKJ maintained about the same performance with an additional layer. None of the NTKs show a noticeable improvement in performance with an additional layer.  

        If the equivalence approximately holds, we would expect to see similar mean RMSE between NN1, NTKA1, NTKB1, and NTKJ1. Similarly, the mean RMSE should be similar between NN2, NTKA2, NTKB2, and NTKJ2. What we observe in Figure \ref{table of RMSE} is that NN1 and NN2 have almost three times the mean RMSE of the corresponding NTKs. NTKB and NTKJ's performance stayed about the same with an additional layer. This is consistent with the assertion made by \citet{simon2022reverse} that adding additional layers in the NTK does not increase kernel expressivity. 
        
        Additionally, we would expect to see the same direction and similar percentage of change in performance with an additional layer to the neural network and the corresponding NTKs. For example, the neural network's mean RMSE in Figure \ref{table of RMSE} decreased by about $7.35\%$ with an additional layer so we would expect to see a similar decrease in RMSE in the corresponding NTK predictors. However, the mean RMSE for NTKJ decreased by $0.098\%$, the mean RMSE for NTKB increased by about $0.013\%$, and the mean RMSE for NTKA increased by about $3.43\%$. None of the NTK variants we tested came close to matching the percentage change in mean RMSE of the NN when we added an additional layer.  

        Furthermore, we notice that the performances of GP1 and GP2 are very close to the performance of NTKB and NTKJ variants in this experiment. Both GP kernels are obtained from a Central Limit Theorem argument that describes the output of an ``infinitely wide'' neural network at initialization (pre-training). Such kernels are not designed to describe trained neural networks. Yet, we see that they have predictive performance on par with NTKB and NTKJ. This should not be possible if the equivalence theorem holds because pre-trained neural networks do not learn anything from the data.
        
        Note that the Laplacian kernel (K1) is completely unrelated to the architecture of the neural networks, yet it achieves a fair predictive performance in this experiment. In fact, the mean RMSE of the Laplacian kernel is closer to that of the neural networks than any of the NTKs.

        These observations suggest the equivalence theorem in \citet{arora2019exact} does not hold well in practice and puts into question whether NTKs adequately address the training process of neural networks.

\section{Conclusion and Discussion}

    Our implementation of 1 layer and 2 layer neural networks and their corresponding NTK predictors does not exhibit the equivalence as claimed by \citet{arora2019exact}. These observations suggest  the assumptions of NTKs and the accompanying equivalence theorem may be too strong to be satisfied in practice.

    We think part of the issue with NTKs is that the Taylor expansion in its derivation requires the lazy training assumption, which is too strong in general. Additionally, an implicit assumption of kernel ridge regression is that the output is linear in some feature space. So informally, NTKs linearize neural networks, twice. 


    \textbf{Acknowledgments}
    
    We acknowledge support from the US Department of Defense [Contract
    No. W52P1J2093009].
    
\newpage 
\section{Appendix}
    \subsection{Hardware Used}
    Intel(R) Xeon(R) W-2295 CPU @ 3.00GHz

    NVIDIA RTX A5000 (24GB)

    125 GiB RAM
\subsection{Description of Code Files}
    \subsubsection{15d\_ntkb1\_par.py}
        The main script that contains both the simulated data generation, neural network training, and implementation of kernel predictors. The outputs include a pickle file (which can then be read later to produce a table of mean RMSEs along with the standard error) and a boxplot of RMSEs. By default, this script will place the output files in the same folder it is located in.

        The outputs of this file should reproduce the figures and tables used in the main body of this paper. Note that this script can take a couple of days to run for 50 trials. It is strongly recommended to time 1 trial first and then estimated how long 50 trials will take before running all 50 trials.  
        
    \subsubsection{parallel\_functions.py}
        Nothing needs to be changed in this file. It contains the parallelized (CPU) functions to compute the kernel matrices. The formulas for the kernels can be found here. 
        
    \subsubsection{read\_pickle.py}
        Input filepaths should be updated in this file for new users. This file reads the pickle file produced by 15d\_ntkb1\_par.py and produces a table (like the one in the main paper).
        
\subsection{Standard Errors table for results}
    In addition to mean RMSE presented in the paper, the table that is produced by our code also contains the standard error of the mean RMSE. This is the output  
    \begin{figure}
        \begin{center}
            \begin{tabular}{ |c|c|c|c| } 
            \hline
            Method & Number of Trials& Sample Mean RMSE & SE of Sample Mean RMSE \\
            \hline
            \hline
            NN (1 layer) & 50 & 0.207151 & 0.0022229 \\ 
            \hline
            NN (2 layers) & 50 & 0.191916  & 0.00202076 \\ 
            \hline
            GP (1 layers)  & 50 & 0.0634271  & 0.000419303\\ 
            \hline
            GP (2 layers) & 50 & 0.0635353  & 0.000420746\\ 
            \hline
            NTKA (1 layers) & 50 & 0.0808606  & 0.000538324\\ 
            \hline
            NTKA (2 layers) & 50 & 0.0836348  & 0.000625299\\ 
            \hline
            NTKJ (1 layers) & 50 & 0.063008  & 0.000412339\\ 
            \hline
            NTKJ (2 layers) & 50 & 0.062946 & 0.000412801\\ 
            \hline
            NTKB (1 layers) & 50 & 0.0629396 & 0.000413537\\ 
            \hline
            NTKB (2 layers) & 50 & 0.0629477 & 0.000413594\\ 
            \hline
            Laplacian (K1) & 50 & 0.103857 & 0.000862339\\ 
            \hline
            \end{tabular}
        \end{center}
        \caption{table of mean RMSE and SE of mean RMSE}
    \end{figure}

    \subsection{Setup for NTKB}
    The following setup can be obtained by setting $\ell = 1$ in Equation (1) in section 2.1 of the main paper and making the adjustments in section 3.1 (see the paragraph on NTKB1 and NTKB2). Namely, the $\frac{1}{n_0}$ scaling factor is removed, but the initial set of weights now have variance $\frac{\sigma_{w,0}^2}{n_0}$ instead of 1. 
    
        \begin{equation}
            z_i^{(1)} 
            = 
            b_i^{(1)} 
            +
            \sum_{j=1}^{n_1} 
            W_{ij}^{(1)}
            \sqrt{\frac{c_{1}}{n_1}} 
            \sigma
            \paren{
                b_j^{(0)} + 
                \sum_{k=1}^{n_0} 
                W_{jk}^{(0)} 
                x_k^{(0)}(\mathbf{x})
            }
        \end{equation}
        where 
            $b_i^{(1)} \sim \mathcal{N}(0,\sigma_{b,1}^2) ,\
            \{b_i^{(0)}\}_i \sim \mathcal{N}(0,\sigma_{b,0}^2) ,\
            \{W_{jk}^{(0)}\}_{jk} \sim \mathcal{N} \paren{0, \frac{\sigma_{w,0}^2}{n_0}} ,\
            \{W_{ij}^{(1)}\}_j \sim \mathcal{N}\paren{0, \sigma_{w,1}^2}$
 
        Notice that we use the indices $i,j,k,q$ in place of $j_\ell$ in our main paper. While $j_\ell$ is convenient for expressing a general formula, having distinguishable letters makes derivations easier to understand. 

    \subsubsection{Partial Derivatives}
        The necessary partial derivatives with respect to parameters:
    
        \begin{align*}
            \frac{\partial z_i^{(1)}(\mathbf{x})}{b_i^{(1)}} &= 1 
            ,\quad
            \frac{z_i^{(1)}(\mathbf{x})}{\partial b_1^{(0)}} = 
            W_{1i}^{(1)}
            \sqrt{\frac{c_{1}}{n_1}}
            \dot{\sigma}(z_1^{(0)}(\mathbf{x})) 
            ,\quad \dots \dots \quad,
            \frac{\partial z_i^{(1)}(\mathbf{x})}{\partial b_{n_1}^{(0)}} 
            =
            W_{in_1}^{(1)} 
            \sqrt{\frac{c_{1}}{n_1}}
            \dot{\sigma}\paren{z_{n_1}^{(0)}(\mathbf{x})} 
        \end{align*}

        \begin{align*}
            \frac{\partial z_i^{(1)}(\mathbf{x})}{\partial W_{11}^{(0)}} 
            =
            W_{i1}^{(1)}
            \sqrt{\frac{c_{1}}{n_1}}
            \dot{\sigma}\paren{z_1^{(0)}(\mathbf{x})}x_1 
            ,\quad \dots \dots \quad,
            \frac{\partial z_i^{(1)}(\mathbf{x})}{\partial W_{1n_0}^{(0)}} = W_{i1}^{(1)} \sqrt{\frac{c_{1}}{n_1}} \dot{\sigma} \paren{z_1^{(0)}(\mathbf{x})}x_{n_0} 
        \end{align*}

        $$\vdots$$
        \begin{align*}  
            \frac{\partial z_i^{(1)}(\mathbf{x})}{\partial W_{n_1 1}^{(0)}} = W_{in_1}^{(1)} \sqrt{\frac{c_{1}}{n_1}} \dot{\sigma} \paren{z_{n_1}^{(0)}(\mathbf{x})}x_{1} 
            ,\quad \dots \dots \quad,
            \frac{\partial z_i^{(1)}(\mathbf{x})}{\partial W_{n_1 n_0}^{(0)}}= W_{in_1}^{(1)} \sqrt{\frac{c_{1}}{n_1}} \dot{\sigma} \paren{z_{n_1}^{(0)}(\mathbf{x})}x_{n_0}
        \end{align*}

        \begin{align*}
            \frac{\partial z_i^{(1)}(\mathbf{x})}{\partial W_{i1}^{(1)}} 
            =
            \sqrt{\frac{c_{1}}{n_1}} \sigma \paren{z_{1}^{(0)}(\mathbf{x})} 
        \end{align*}

    \subsubsection{NTK Expression}
        \begin{align*}
            &\inner
            {
                \frac
                {\partial z_i^{(1)}(\mathbf{x})}
                {\partial \pmb{\theta}}
            }
            {
                \frac
                {\partial z_i^{(1)}(\mbf{x'})}
                {\partial \pmb{\theta}}
            } 
            = 1 
            + 
            c_{1}
            \frac{1}{n_1} 
            \sum_{k=1}^{n_1} 
            \sigma \paren{z_1^{(0)}(\mathbf{x})}
            \sigma \paren{z_1^{(0)}(\mbf{x'})} \\ 
            &+
            c_{1}
            \frac{1}{n_1} 
            \sum_{k=1}^{n_1}
            \paren{W_{ik}^{(1)}}^2
            \dot{\sigma}\paren{z_k^{(0)}(\mathbf{x})}
            \dot{\sigma}\paren{z_k^{(0)}(\mbf{x'})}
            + 
            c_{1}
            \frac{1}{n_1} 
            \sum_{j=1}^{n_1} 
            \sum_{k=1}^{n_0}
            \paren{W_{ij}^{(1)}}^2 
            \dot{\sigma}\paren{z_j^{(0)}(\mathbf{x})}
            x_k
            \dot{\sigma}\paren{(z_j^{(0)}(\mbf{x'})}
            y_k 
            \numberthis\\
        \end{align*}
        Using the Weak Law of Large Numbers, the summations in the above terms converge in probability to the following expected values:
        \begin{align*}
            \inner
            {
                \frac
                {\partial z_i^{(1)}(\mathbf{x})}
                {\partial \pmb{\theta}}
            }
            {
                \frac
                {\partial z_i^{(1)}(\mbf{x'})}
                {\partial \pmb{\theta}}
            } 
            &
            \convProb
            1 
            +
            c_{1}
            \ex
            \bracket{
                \sigma\paren{z_1^{(0)}(\mathbf{x})}
                \sigma\paren{z_1^{(0)}(\mbf{x'}))}
            } 
            +
            c_{1}
            \ex
            \bracket{
                \paren{
                    W_{ik}^{(1)}
                }^2 
            }
            \ex 
            \bracket{
            \dot{\sigma}\paren{z_k^{(0)}(\mathbf{x})}
            \dot{\sigma}\paren{(z_k^{(0)}(\mbf{x'}))}} \\
            &+ 
            c_{1} 
            \sum_{k=1}^{n_0}
            x_k
            y_k
            \ex
            \bracket{
                \paren{W_{ij}^{(1)}}^2
            }
            \ex 
            \bracket{
                \dot{\sigma}\paren{z_j^{(0)}(\mathbf{x})}
                \dot{\sigma}\paren{z_j^{(0)}(\mbf{x'})}
            } 
            \numberthis
        \end{align*}
        
        \begin{align*}
            \inner{\frac{\partial z_i^{(1)}(\mathbf{x})}{\partial \pmb{\theta}}}
            {\frac{\partial z_i^{(1)}(\mbf{x'})}{\partial \pmb{\theta}}} 
            &= 
            1 
            +
            c_{1} 
            \ex 
            \bracket{
                \sigma\paren{z_1^{(0)}(\mathbf{x})}
                \sigma\paren{z_1^{(0)}(\mbf{x'})}
            }  
            + 
            c_{1}
            \sigma_{w,1}^2
            \ex 
            \bracket{
                \dot{\sigma}\paren{z_k^{(0)}(\mathbf{x})}
                \dot{\sigma}\paren{z_k^{(0)}(\mbf{x'})}
            }  \\
            &+ 
            c_{1} 
            \sigma_{w,1}^2
            \ex 
            \bracket{
                \dot{\sigma}\paren{z_j^{(0)}(\mathbf{x})}
                \dot{\sigma}\paren{z_j^{(0)}(\mbf{x'})}
            } 
            \inner{\mathbf{x}}{\mbf{x'}}\\
            &= 
            1 
            +
            c_{1} 
            \frac{1}{2\pi}
            \sqrt{
                \paren{
                    \frac{\sigma_{w,0}^2}{n_0}
                    \norm{\mathbf{x}}^2 
                    + 
                    \sigma_{b,0}^2
                }
                \paren{
                    \frac{\sigma_{w,0}^2}{n_0}
                    \norm{\mbf{x'}}^2 
                    +
                    \sigma_{b,0}^2
                }
            }
            \paren{
                \sin(\delta^{(0)}) 
                + 
                \cos(\delta^{(0)})(\pi - \delta^{(0)})
            } \\
            &+
            c_{1} 
            \sigma_{w,1}^2
            \frac{1}{2\pi} 
            \paren{\pi-\delta^{(0)}} 
            + 
            c_{1}
            \sigma_{w,1}^2
            \frac{1}{2\pi}
            \inner{\mathbf{x}}{\mbf{x'}}
            \paren{\pi - \delta^{(0)}}
            \numberthis
        \end{align*}
    where 
    \begin{align*}
        \delta^{(0)}
        &= 
        \arccos
        \paren{
            \frac{
                \paren{
                    \frac{\sigma_{w,0}^2}{n_0}
                    \inner{\mathbf{x}}{\mbf{x'}} 
                    +
                    \sigma_{b,0}^2
                }
            }
            {
                \sqrt{
                    \paren{
                        \frac{\sigma_{w,0}^2}{n_0}
                        \norm{\mathbf{x}}^2
                        +
                        \sigma_{b,0}^2
                    }
                    \paren{
                        \frac{\sigma_{w,0}^2}{n_0}
                        \norm{\mbf{x'}}^2
                        +
                        \sigma_{b,0}^2
                    }
                }
            }
        }
        \numberthis
    \end{align*}
    and details for computing the expectations can be found in \cite{cho2009kernel}  
     
\subsection{Derivations for NTKB2}
    The following setup can be obtained by setting $\ell = 2$ in Equation (1) in section 2.1 of the main paper and making the adjustments in section 3.1 (see the paragraph on NTKB1 and NTKB2). Namely, the $\frac{1}{n_0}$ scaling factor is removed, but the initial set of weights now have variance $\frac{\sigma_{w,0}^2}{n_0}$ instead of 1. 

    \begin{equation}
        z_{q}^{(2)}(\mathbf{x}) 
        = 
        b_{{q}}^{(2)}
        + 
        \sum_{i=1}^{n_2}
        W_{{q} i}^{(2)}
        \sqrt{
            \frac{c_2}{n_2}
        }
        \sigma
        \paren{
            b_i^{(1)} 
            +
            \sum_{j=1}^{n_1}
            W_{ij}^{(1)}
            \sqrt{
                \frac{c_1}{n_1}
            }
            \sigma
            \paren{
                b_j^{(0)}
                +
                \sum_{k=1}^{n_0}
                W_{jk}^{(0)}
                x_k^{(0)}(\mathbf{x})
            }
        }
    \end{equation}
    
    where
        $b_{{q}}^{(2)}  \sim \mathcal{N}(0, \sigma_{b,2}^2) ,\,
        \{b_i^{(1)}\}_{i=1}^{n_2} \sim \mathcal{N}(0,\sigma_{b,1}^2) ,\,
        \{b_i^{(0)}\}_{j=1}^{n_1} \sim \mathcal{N}(0,\sigma_{b,0}^2)
        \{W_{jk}^{(0)}\}_{jk} \sim \mathcal{N} \paren{0, \frac{\sigma_{w,0}^2}{n_0}},\,$

        $
        \{W_{ij}^{(1)}\}_{ij} \sim \mathcal{N}\paren{0, \sigma_{w,1}^2} ,\,
        \{W_{{q} i}^{(2)}\}_i \sim \mathcal{N}\paren{0, \sigma_{w,2}^2}$
   
    In a similar fashion to NTKB1: we compute the partial derivatives, compute the inner product, gather terms, and apply the Weak Law of Large numbers to obtain:
    
    \begin{align*}
        &\inner
        {
            \frac
            {
                \partial z_{q}^{(2)}(\mathbf{x})
            }
            {
                \partial \pmb{\theta}
            }   
        }
        {
            \frac
            {
                \partial z_{q}^{(2)}(\mbf{x'})
            }
            {
                \partial \pmb{\theta}
            }   
        }\\
        &= 1  
        +
        c_2 
        \sigma_{w,2}^2
        \ex 
        \bracket{
            \dot{\sigma}
            \paren{
                z_{i}^{(1)}(\mathbf{x})
            }
            \dot{\sigma}
            \paren{
                z_{i}^{(1)}(\mbf{x'})
            }
        } 
        +
        c_1
        c_2
        \sigma_{w,2}^2
        \sigma_{w,1}^2
        \ex 
        \bracket{
            \dot{\sigma}
            \paren{
                z_i^{(1)}(\mathbf{x})
            }
            \dot{\sigma}
            \paren{
                z_i^{(1)}(\mbf{x'})
            }
        }
        \ex 
        \bracket{
            \dot{\sigma}
            \paren{
                z_j^{(0)}(\mathbf{x})
            }
            \dot{\sigma}
            \paren{
                z_j^{(0)}(\mbf{x'})
            }
        } \\
        &+
        c_1 
        c_2
        \sigma_{w,1}^2
        \sigma_{w,2}^2
        \inner{\mathbf{x}}{\mbf{x'}}
        \ex 
        \bracket{
            \dot{\sigma}
            \paren{z_i^{(1)}(\mathbf{x})}
            \dot{\sigma}
            \paren{z_i^{(1)}(\mbf{x'})}
        }
        \ex 
        \bracket{
            \dot{\sigma}
            \paren{z_j^{(0)}(\mathbf{x})}
            \dot{\sigma}
            \paren{z_j^{(0)}(\mbf{x'})}
        } \\
        &+ 
        c_1
        c_2
        \sigma_{w,2}^2
        \ex 
        \bracket{
            \dot{\sigma}
            \paren{z_i^{(1)}(\mathbf{x})}
            \dot{\sigma}
            \paren{z_i^{(1)}(\mbf{x'})}
        }
        \ex
        \bracket{
            \sigma 
            \paren{z_j^{(0)}(\mathbf{x})}
            \sigma
            \paren{z_j^{(0)}(\mbf{x'})}
        } 
        +
        c_2
        \ex 
        \bracket{
            \sigma
            \paren{z_i^{(1)}(\mathbf{x})}
            \sigma
            \paren{z_i^{(1)}(\mbf{x'})}
        }
        \numberthis
    \end{align*}

    where
    \begin{align*}
        \ex
        \bracket{
            \dot{\sigma}
            \paren{z_i^{(1)}(\mathbf{x})}
            \dot{\sigma}
            \paren{z_i^{(1)}(\mbf{x'})}
        }
        &= 
        \frac{1}{2 \pi}
        \paren{\pi - \delta^{(1)}} 
        \quad,\quad
        \ex 
        \bracket{
            \dot{\sigma}\paren{z_j^{(0)}(\mathbf{x})}
            \dot{\sigma}\paren{(z_j^{(0)}(\mbf{x'})}
        }
        = 
        \frac{1}{2 \pi}
        \paren{\pi - \delta^{(0)}} \\
        \ex 
        \bracket{
            \sigma 
            \paren{z_j^{(0)}(\mathbf{x})}
            \sigma 
            \paren{z_j^{(0)}(\mathbf{x})}
        }
        &=
        \frac{1}{2\pi}
        \sqrt{
            K^{(0)}(\mathbf{x}, \mathbf{x})
            K^{(0)}(\mbf{x'}, \mbf{x'})
        }
        \paren{
            \sin(\delta^{(0)}) 
            + 
            \cos(\delta^{(0)})(\pi - \delta^{(0)})
        }  \\
        \ex 
        \bracket{
            \sigma 
            \paren{z_j^{(1)}(\mathbf{x})}
            \sigma 
            \paren{z_j^{(1)}(\mathbf{x})}
        }
        &=
        \frac{1}{2\pi}
        \sqrt{
            K^{(1)}(\mathbf{x}, \mathbf{x})
            K^{(1)}(\mbf{x'}, \mbf{x'})
        }
        \paren{
            \sin(\delta^{(1)}) 
            + 
            \cos(\delta^{(1)})(\pi - \delta^{(1)})
        } \\
        K^{(0)}(\mathbf{x}, \mbf{x'})
        &= 
        \frac{\sigma_{w,0}^2}{n_0}
        \inner{\mathbf{x}}{\mbf{x'}}
        + 
        \sigma_{b,0}^2 \\
        K^{(1)}(\mathbf{x}, \mbf{x'})
        &= 
        \sigma_{b,1}^2 
        +
        \frac{c_{1} \sigma_{w,1}^2}{2\pi}
        \sqrt{
            K^{(0)}(\mathbf{x}, \mathbf{x}) 
            K^{(0)}(\mbf{x'}, \mbf{x'})
        }
        \paren{
            \sin(\delta^{(0)}) 
            +
            \cos(\delta^{(0)})(\pi - \delta^{(0)})
        } \\
        \delta^{(0)} 
        &= 
        \cos^{-1}
        \paren{
            \frac{
                K^{(0)}(\mathbf{x}, \mbf{x'})
            }
            {
                \sqrt{
                    K^{(0)}(\mathbf{x}, \mathbf{x})
                    K^{(0)}(\mbf{x'}, \mbf{x'})
                }   
            }
        } 
        \quad , \quad
        \delta^{(1)} 
        = 
        \cos^{-1}
        \paren{
            \frac{
                K^{(1)}(\mathbf{x}, \mbf{x'})
            }
            {
                \sqrt{
                    K^{(1)}(\mathbf{x}, \mathbf{x})
                    K^{(1)}(\mbf{x'}, \mbf{x'})
                }   
            }
        }
        \numberthis
    \end{align*}

    the computation for the expectations can be found in \citet{cho2009kernel}, and
    $K^{(0)}$ and $K^{(1)}$ can be found in \citet{neal1996priors} and \citet{lee2017deep}.

\newpage

\bibliographystyle{rusnat}
\bibliography{refs.bib}

\end{document}